\documentclass[runningheads]{llncs}
\usepackage[T1]{fontenc}
\usepackage{graphicx}
\usepackage{amsmath}
\usepackage{booktabs}
\usepackage{rotating}
\usepackage{float}
\usepackage{placeins}
\usepackage{multirow}
\usepackage{subcaption}

\usepackage[most]{tcolorbox}
\usepackage{capt-of}
\usepackage{needspace}
\usepackage{adjustbox}

\tcbset{
  compactprompt/.style={
    enhanced,
    sharp corners,
    boxrule=0.3pt,
    colback=white,
    colframe=black,
    left=1mm,
    right=1mm,
    top=0.5mm,
    bottom=0.5mm,
    boxsep=0.5mm,
    before skip=0pt,
    after skip=0pt,
    fontupper=\scriptsize\linespread{0.90}\selectfont,
    before upper={\setlength{\parskip}{0pt}\setlength{\parindent}{0pt}}
  }
}

\newtcolorbox{promptbox}[1][]{%
  compactprompt,
  width=\linewidth,
  #1
}

\begin{document}
\title{Mitigating Scoring Bias in LLM-as-a-Judge via Random Number Generation}
\titlerunning{Mitigating Scoring Bias via Random Number Generation}
%
\author{
    Yuma Asato \and 
    Kiyoaki Shirai \and 
    Natthawut Kertkeidkachorn 
}
\authorrunning{Asato et al.}
\institute{
Japan Advanced Institute of Science and Technology \\
\email{\{yumaasato, kshirai, natt\}@jaist.ac.jp}}

\maketitle             
\begin{abstract}
Large Language Models (LLMs) are often used as evaluators of text quality, known as LLM-as-a-Judge, which can outperform conventional automatic evaluation metrics that rely on reference texts. 
However, LLM evaluators tend to generate particular scores regardless of the context of the evaluated text, which is known as scoring bias.
This study proposes a novel method to mitigate this scoring bias. 
An LLM is instructed to randomly generate number tokens, and the latent numerical bias of the LLM is identified by measuring the deviation of the observed distribution of numbers from the uniform distribution.
A definition of a downstream task, for which an LLM evaluator is used, is added to the prompts for random number generation to measure task-specific latent number bias. 
In the evaluation by an LLM, the token generation probabilities for a given input are rectified considering the LLM's latent number bias.
Results of the experiment on four different tasks, evaluation of LLM alignment, evaluation of summarization, Semantic Textual Similarity, and Semantic Textual Relatedness, demonstrate that our proposed method outperforms the baselines, including an LLM without debiasing and previous calibration methods. 
In addition, it is confirmed that scoring bias varies across LLMs, tasks, and score ranges, indicating the importance of measuring latent number bias as the case may be.
\end{abstract}

\section{Introduction} 
\label{sec:intro}
Large Language Models (LLMs) have achieved strong performance in a wide range of natural language processing tasks, including machine translation and summarization.
In recent years, LLMs have also been used to evaluate the quality of texts, a paradigm known as LLM-as-a-Judge.
Conventional automatic evaluation criteria for natural language generation (NLG) have mainly relied on methods that measure the similarity between a generated text and a reference text, such as BLEU~\cite{papineni-etal-2002-bleu}, ROUGE~\cite{lin-2004-rouge}, and BERTScore~\cite{Zhang*2020BERTScore:}. 
In contrast, using LLMs as evaluators is known not only to reduce the human cost of preparing a substantial number of reference texts, but also to be sometimes more appropriate than reference-based automatic evaluation~\cite{liu-etal-2023-g}.
However, it is well known that LLM evaluators exhibit various biases.
Especially, scoring bias can degrade the performance of the LLM's evaluation.
In the context of LLM-as-a-Judge, scoring bias is defined as the tendency to exhibit a preference for specific numerical values regardless of the context of the evaluated text~\cite{sato2026exploringeffectsalignmentnumerical}.
\par
Besides, several studies investigated LLM's ability for random generation~\cite{zhao2026largelanguagemodelsbad,west2025base,gu2026illusionstochasticityllms,hopkins2023can}.
The ``ability for random generation'' refers to the behavior of LLMs to generate genuinely random outputs in tasks that require unpredictability, such as random number generation.
These studies demonstrated that LLMs do not exhibit sufficient randomness; in particular, instruction-tuned LLMs often fail to generate outputs that follow a uniform distribution in random number generation.
The lack of randomness suggests an undesirable preference for specific numbers, which could lead to scoring bias in LLM-as-a-Judge.
However, previous studies have overlooked an approach to identifying scoring bias from the perspective of ability for random generation of LLMs and correcting it.
\par
This paper proposes a novel method for mitigating scoring bias by considering deviations from genuine random generation in LLMs to improve the performance of LLM evaluators. 
Random number generation is performed to measure inherent bias in predicting the scores of a target text, i.e., the tendency for numbers to be generated more or less frequently regardless of the input. 
In addition, supposing that the inherent bias of LLMs may differ across downstream evaluation tasks, a task definition is provided to the prompt for random number generation. 
Then, the probabilities of token generation by an LLM are modified so that the probabilities of positively (or negatively) biased numbers are reduced (or increased). 
The effectiveness of the proposed method is evaluated through experiments on four different evaluation tasks. 
The results indicate that our method outperforms the previous debiasing methods.
\par
Our main contributions are summarized as follows.\footnote{Codes will be available upon acceptance.}
\begin{itemize}
    \item We propose a debiasing method that estimates the LLM's inherent scoring bias and mitigates it during evaluation by the LLM.
    \item We propose a task-dependent debiasing method that mitigates the inherent scoring bias in individual downstream tasks of LLM-as-a-Judge.
    \item We demonstrate the effectiveness of the proposed method through comprehensive experiments, including four evaluation tasks, five LLMs, and a comparison with representative previous calibration methods.
\end{itemize}

\section{Related Work}
\label{sec:related}
\subsection{LLM-as-a-Judge}
As described in section~\ref{sec:intro}, LLM-as-a-Judge has been widely used in recent years.
Unlike traditional reference-based metrics (e.g., BLEU~\cite{papineni-etal-2002-bleu}, ROUGE~\cite{lin-2004-rouge}, and BERTScore~\cite{Zhang*2020BERTScore:}), LLM evaluators can evaluate the quality of texts without reference texts.
Fu et al. proposed GPTScore, a framework that enables generative pre-trained models to evaluate text quality across multiple aspects and demonstrated its effectiveness~\cite{fu-etal-2024-gptscore}.
Liu et al. proposed G-Eval, a framework to assess the quality of NLG tasks such as text summarization and response generation in a dialog system~\cite{liu-etal-2023-g}.
G-Eval outperformed conventional reference-based and reference-free automatic evaluation metrics in terms of Spearman correlation with human judgments.
Prometheus 2 is an open-source LLM specializing in evaluation, which is fine-tuned using feedback datasets for two evaluation formats: direct evaluation by generating a score and relative evaluation via pairwise ranking~\cite{kim-etal-2024-prometheus}.
It demonstrated evaluation performance comparable to GPT-4~\cite{achiam2023gpt}.
Zheng et al. introduced MT-Bench, a multi-turn conversation benchmark, and Chatbot Arena, a crowdsourcing benchmark platform~\cite{zheng2023judging}.
They have shown that strong LLMs achieve a high agreement rate with human preferences, while also exhibiting biases such as position bias and verbosity bias in LLM-as-a-Judge.
\subsection{Bias of LLM Evaluator}
Several types of scoring bias of LLM evaluators are reported.
Sato et al. pointed out the problem of scoring bias, referred to as ``numerical bias'' in their paper, and investigated three approaches to mitigate it: modifying the temperature, Distribution Calibration, and adjusting the score range~\cite{sato2026exploringeffectsalignmentnumerical}.
Li et al. reported that the scores produced by LLM evaluators vary substantially in response to minor prompt changes, including the order of the scoring rubric, the notation of score ID, and the attachment of reference answer score~\cite{10.1007/978-981-92-0372-7_2}.
Fujinuma proposed a method to mitigate a score range bias, a phenomenon in which LLMs' outputs are highly sensitive to pre-defined score ranges~\cite{fujinuma2026contrastivedecodingmitigatesscore}. 
Similar to Sato's work~\cite{sato2026exploringeffectsalignmentnumerical}, this study focuses on scoring bias, LLM's inherent tendency to frequently generate specific scores irrespective of the input, and proposes a method to mitigate it.

\subsection{Calibration Method for LLMs}
Numerous attempts have been devoted to bias mitigation or calibration for LLMs.
This subsection introduces two representative calibration methods that are compared with the proposed method in our experiments.
Contextual Calibration~\cite{pmlr-v139-zhao21c} is the calibration method for LLMs.
It measures bias in classification labels derived from the token generation probability distribution for context-free inputs such as \texttt{N/A}, and corrects the label generation probabilities at inference based on this bias.
Sato et al. mitigated scoring bias by adopting the calibration method proposed by Jiang et al.~\cite{jiang-etal-2023-generative}, which they referred to it as Distribution Calibration (DC)~\cite{sato2026exploringeffectsalignmentnumerical}.
It calibrates the label marginal probability $p(y)$ via Monte-Carlo sampling, while estimating $q(y)$ via fitting a Beta distribution to the ground-truth scores using maximum likelihood estimation.
Finally, the prediction score is reweighted by the ratio $q(y)/p(y)$.

\section{Proposed Method}
\label{sec:method}
\begin{figure}[t]
 \centering
 \includegraphics[width=\columnwidth]{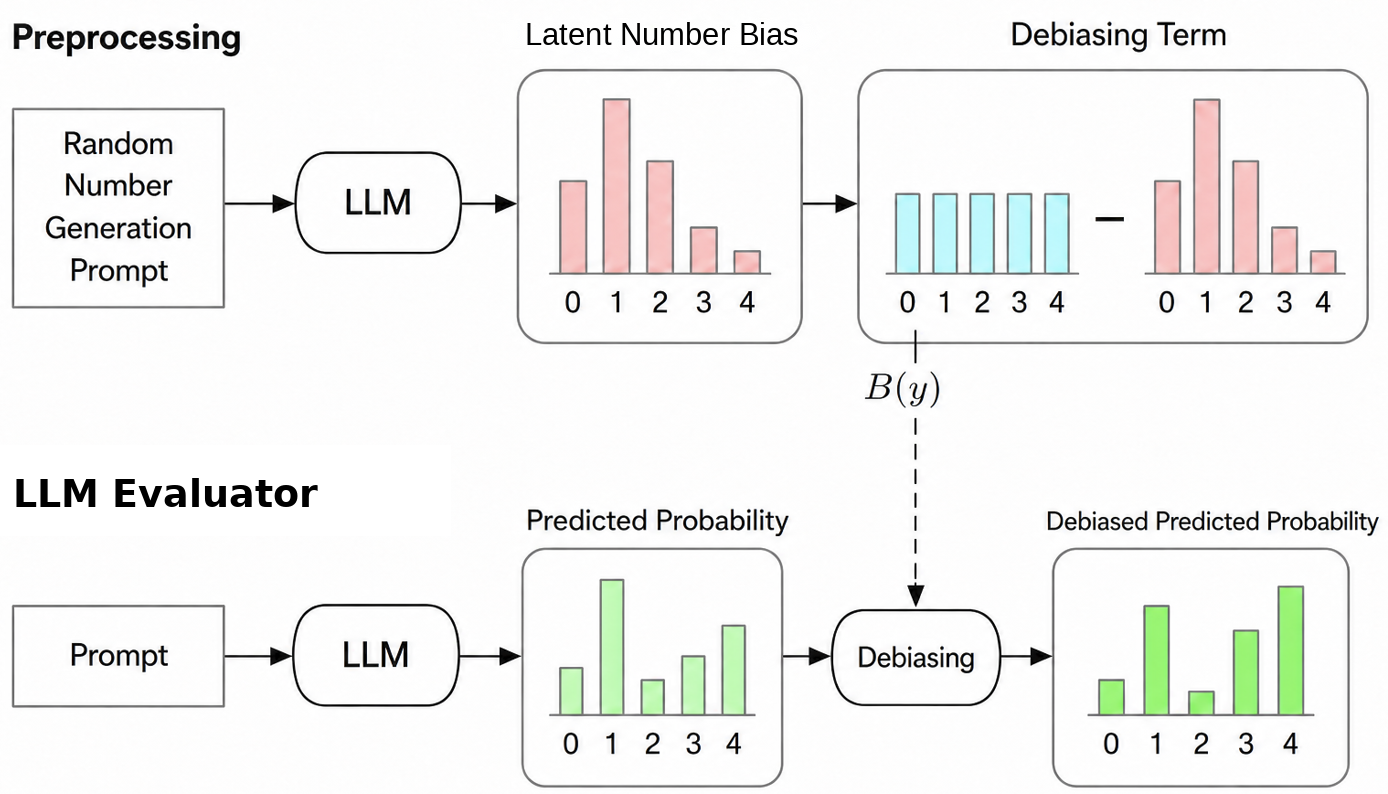}
 \caption{Overview of our proposed method.}
 \label{fig:overview}
\end{figure}

Let us suppose that an LLM is used as an evaluator by giving a prompt that forces it to assess the quality of an input text and generate a score as an integer in $\mathcal{Y}=\{0,\ldots,N\}$.
The goal of this study is to mitigate the inherent bias of an LLM in generating numerical tokens.\par
Figure~\ref{fig:overview} shows an overview of the proposed method.
In the preprocessing stage, several prompts are given to an LLM to randomly generate a number, then a probabilistic distribution over generating numerical tokens in the set $\mathcal{Y}$ is observed.
This distribution often deviates from a uniform distribution, indicating the presence of bias in the LLM.
Hereafter, this probability distribution is referred to as ``latent number bias.''
Then, the debiasing term $B(y)$ is calculated by subtracting the latent number bias from the uniform distribution.
In the evaluation stage, when an LLM is used as an evaluator, the predicted probability distribution over the scores is revised by adding the debiasing terms to the logits of the number tokens. 
The evaluation score is then determined based on this debiased probability distribution.

\subsection{Measurement of Latent Number Bias}
\label{subsec:calculating_latent_bias}
To measure the latent number bias of an LLM, prompts instructing the LLM to randomly generate numbers are provided.
First, 100 prompts for instructing random number generation are generated using ChatGPT-o3 with the prompt shown in Figure~\ref{fig:random_prompt_generation}.
The variables \texttt{\{\{min\_score\}\}} and \texttt{\{\{max\_score\}\}} denote the minimum and maximum scores defined in an LLM evaluator, i.e., 0 and N, respectively.\par
Next, only reliable prompts are retained to measure the latent number bias. 
Specifically, the perplexity of each prompt is computed, then the top $T_p$ prompts with the lowest perplexity are chosen.
$T_p$ is intuitively set to 10 in this study.

\begin{figure}[tbp]
\centering
\fbox{%
  \begin{minipage}{0.91\linewidth}
  Generate 100 prompts that have the same meaning as the following prompt.\\
  Please generate a random integer between \texttt{\{\{min\_score\}\}} and \texttt{\{\{max\_score\}\}}.
  \end{minipage}%
}
\caption{Meta-prompt to generate prompts for random number generation.}
\label{fig:random_prompt_generation}
\medskip
\end{figure}

Probability distributions over numerical tokens are elicited from an LLM via random number generation with the chosen prompts.
Let $y$ denote a numerical token, and $p(y|x_i)$ denote the probability that the LLM generates $y$ given the $i$-th prompt $x_i$.
The average probability distribution of each numerical token is computed by Equation (\ref{eq:random_potential_bias}).
 
\begin{equation}
  \label{eq:random_potential_bias}
  P_{\mathrm{random}}(y) 
  = 
  \frac{1}{T_p} 
  \sum_{i=1}^{T_p}
  \frac{p(y \mid x_i)}
{\sum_{y \in \mathcal{Y}} p(y \mid x_i)}
\end{equation}
In addition, to avoid $P_{\mathrm{random}}(y)$ being zero, $P_{\mathrm{random}}(y)$ is set to the minimum probability $p_{\mathrm{min}}$, which is defined as $10^{-12}$, if it is zero or less than $p_{\mathrm{min}}$.
The resulting distribution $P_{\mathrm{random}}(y)$ is utilized as the LLMs' latent number bias.\par
Two settings are considered for measuring the latent numerical bias. One is the simple random setting, denoted as ``Deb-R'', where the prompts for random number generation are used in their original form. 
The other is the task-conditioned random setting, denoted as ``Deb-T''. 
It is expected that the latent number bias of an LLM may vary when used to evaluate different downstream tasks.
In Deb-T, simulating the situation where the LLM is used for a downstream task, the instruction for task evaluation is attached before the prompt.
Specifically, the prompt is formulated as
\begin{center}
  \texttt{\{\{task instruction\}\}} \texttt{\{\{random number generation instruction\}\}},
\end{center}
where \texttt{\{\{task instruction\}\}} is the instruction to the LLM that is used as an evaluator for a specific task.

\subsection{Estimation of Debiasing Term}
\label{subsec:debiasing_term}
When an LLM is prompted to generate random numbers, the resulting probability distribution is expected to be uniform; however, in practice, it is not.
This is supported by previous work~\cite{coronadoblzquez2025deterministicprobabilisticpsychologyllms,west2025base}.
The deviation from the uniform distribution is used to quantify the strength of the bias for each numerical token.
This is referred to as the debiasing term \(B(y)\), which is defined as follows.
\begin{equation}
    \label{eq:debiasing_term}
    B(y) = \log{U(y)} - \log{P_{\text{random}}(y)}
\end{equation}

\(U(y)\) denotes the uniform distribution over numerical tokens, i.e., \(\forall y: U(y) = \frac{1}{N + 1}\).
The \(B(y)\) becomes smaller for numerical tokens that are intrinsically more likely to be generated, and greater for less likely generated numerical tokens.

\subsection{Debiasing in Scoring-Based Evaluation}
When evaluating samples in a downstream task with an LLM evaluator, the generation probabilities of numerical tokens are corrected as follows.
\begin{equation}
    \label{eq:debiasing_numeric_token}
    \tilde{P}(y) = \text{softmax}(l(y)+ \lambda \cdot B(y))
\end{equation}
\(l(y)\) denotes the logit corresponding to the numerical token \(y\), and \(\lambda\) is a hyperparameter that controls the strength of the correction.
In our experiment, \(\lambda\) is optimized using the development set.

\section{Evaluation}
\label{sec:evaluation}
\subsection{Task and Dataset}
\label{subsec:task_and_dataset}
Four tasks of LLM-as-a-Judge are employed for experiments to evaluate the proposed debiasing method.
The first two tasks are the evaluation of automatically generated texts. LLM alignment (LLM-Alignment, in short) is a task for evaluating the quality of texts generated by LLMs. HelpSteer2~\cite{wang2024helpsteer} is a dataset for this task, consisting of a prompt, a text generated from the prompt, and a score for the text on a five-point Likert scale from 0 to 4 across five points of view: helpfulness, correctness, coherence, complexity, and verbosity. 
The evaluation for text summarization (Summ) is a task to evaluate summarization systems. 
SummEval~\cite{fabbri-etal-2021-summeval} is a benchmark that comprises an original article in CNN and DailyMail~\cite{hermann2015teaching}, its corresponding summary, and a score on a five-point Likert scale (1--5) for coherence, consistency, fluency, and relevance.\footnote{Actually, an average score of multiple annotators, i.e., a continuous value, is associated with
each sample.}
In this experiment, LLM evaluators are used to predict the helpfulness score for the LLM alignment task and the coherence score in the summarization evaluation task. 
As HelpSteer2 and SummEval do not provide a test set, five-fold cross-validation is conducted, where the dataset is divided into a development and test set at 1:4.\par
The other two tasks are the measurement of scores between two sentences. Semantic Textual Similarity (STS) is a task to predict the similarity between two sentences.
STS-B~\cite{wang-etal-2018-glue} is used in this experiment, which is a benchmark for the STS task.
Similarly, Semantic Textual Relatedness (STR) is a task to predict the extent to which two sentences are semantically related.
SemRel2024~\cite{ousidhoum-etal-2024-semrel2024} is a dataset consisting of sentence pairs and their human-annotated scores.
Both STS-B and SemRel2024 assign continuous scores from 0 to 5 to each sentence pair.
Accordingly, LLMs evaluate the similarity or semantic relatedness of two sentences and predict a score between 0 and 5.
The original split of a test and development set is retained.\par
The number of samples of these datasets is shown in Table~\ref{tab:dataset_statistics}.
The development set is used to optimize the hyperparameter $\lambda$ in Equation (\ref{eq:debiasing_numeric_token}).
The best value is selected from a range of candidates from 0.1 to 1.0 in increments of 0.1.

\begin{table}[htbp]
  \centering
  \caption{Statistics of datasets.}
  \label{tab:dataset_statistics}
  \setlength{\tabcolsep}{6pt}
  \begin{tabular}{l|rr}
    \hline
    Dataset & Dev. & Test \\
    \hline
    HelpSteer2 & 208 & 830 \\ 
    SummEval   & 320 & 1,280 \\ 
    STS-B      & 1,500 & 1,379 \\
    SemRel2024 & 250 & 2,600 \\
    \hline
  \end{tabular}
\end{table}

\subsection{Experiment Settings}
\noindent \textbf{Score Prediction}~~
Following a prior study~\cite{liu-etal-2023-g}, the final LLM's prediction score is not the numerical token with the highest probability, but the weighted sum of the numeric values, where the weights are the generation probabilities of each numerical token.
That is, the score predicted by an LLM, $s_{\text{llm}}$, is computed as follows, 
\begin{equation}
\label{eq:prediction_score}
s_{\text{llm}}
= \sum_{y \in \mathcal{Y}} \tilde{p}(y) \cdot y,
\end{equation}
where $\tilde{p}(y)$ is the generation probability of the token $y$.
A zero-shot prompt is given to the LLMs.
Appendix~\ref{sec:prompt} shows the prompts for each of the four tasks.
The temperature parameter is set to 0.2 in all experiments.

\medskip
\noindent \textbf{LLMs}~~
Five open-source instruction-tuned LLMs are used as evaluators.
Specifically, Llama-3.1-8B-\allowbreak{}Instruct (Llama)\allowbreak{} \cite{L3.1-8B-it},
Qwen2.5-14B-\allowbreak{}Instruct (Qwen)\allowbreak{} \cite{qwen2.5},
Gemma-2-9B-\allowbreak{}Instruct (Gemma)\allowbreak{} \cite{G2-9B-it},
Prometheus-7B-\allowbreak{}v2.0 (Prometheus)\allowbreak{} \cite{kim-etal-2024-prometheus},
and Mistral-7B-\allowbreak{}Instruct-v0.3 (Mistral)\allowbreak{} \cite{M0.3-7B-it} are used.

\medskip
\noindent \textbf{Baseline}~~
Our proposed methods are compared with three baselines: vanilla, Contextual Calibration (CC)~\cite{pmlr-v139-zhao21c}, and Distribution Calibration (DC)~\cite{sato2026exploringeffectsalignmentnumerical}. 
Vanilla is an LLM evaluator without debiasing.

\medskip
\noindent \textbf{Metrics}~~
LLM evaluators are assessed using a variety of statistical metrics, including Spearman's rank correlation coefficient, Pearson correlation coefficient, and mean squared error (MSE) between the predicted and reference scores.
In addition, we define Score Distribution Divergence (SDD) to evaluate the precision of a predicted probability distribution over numerical tokens rather than a predicted score.
SDD is the metric that measures the closeness between the predicted and ground-truth probability distribution.
Let $x$ be a sample in the test set, and $y$ be a numerical token.
$P_x$ is the predicted probability distribution of $p_x(y)$ over $y \in \mathcal{Y}$ when $x$ is given as input, while $Q_x$ is the ground-truth probability distribution.
SDD is the average of the Kullback-Leibler divergence between $P_x$ and $Q_x$ as defined as follows,
\begin{equation}
    \label{eq:sdd}
\mathrm{SDD}
=
\frac{1}{|D|}
\sum_{x \in D}
D_{\mathrm{KL}}(Q_x || P_x)
=
\frac{1}{|D|}
\sum_{x \in D}
\sum_{y \in \mathcal{Y}}
q_{x}(y)
\log
\frac{q_{x}(y)}{p_{x}(y)},
\end{equation}
where $D$ is the test set.
For the dataset HelpSteer2, where a discrete value is annotated as a score, $q_{x}(y)$ is set to 1 for the ground-truth label and 0 otherwise.
In contrast, for the SummEval, STS-B, and SemRel2024 datasets, where a continuous value is annotated, a probability is distributed to the two near tokens in proportion to their proximity.
For example, when the ground-truth score is $3.7$, $q_x(3) = 0.3$ and $q_x(4) = 0.7$, while zero probability is assigned to other tokens.
Lower SDD indicates better performance.

\subsection{Results}
\label{subsec:experiment_results}

\begin{table}[tb]
  \centering
  \setlength{\tabcolsep}{2.5pt}
  \caption{Results of the LLM alignment and summarization tasks. (a) Spearman, (b) Pearson, (c) MSE, and (d) SDD.}
  \label{tab:result-llla-summ}
  \begin{tabular}{clcccccccccc}
    \toprule
    & \textbf{LLM}
    & \multicolumn{5}{c}{\textbf{LLM-Align}} & \multicolumn{5}{c}{\textbf{Summ}} \\
    \cmidrule(lr){3-7}
    \cmidrule(lr){8-12}
    & & Van. & CC & DC & Deb-R & Deb-T & Van. & CC & DC & Deb-R & Deb-T \\
    \midrule
    (a)
    & Llama
    & 13.3 & 11.4 & \textbf{15.9} & 14.9 & 14.9
    & 13.6 & 12.2 & 13.8 & 13.0 & \textbf{18.0} \\
    $\uparrow$
    & Qwen
    & 20.8 & \textbf{24.0} & 11.8 & 23.6 & 23.6
    & 46.4 & 45.3 & 31.1 & 47.6 & \textbf{49.0} \\
    & Gemma
    & 14.0 & \textbf{20.3} & 15.6 & 15.7 & 13.6
    & 31.6 & \textbf{42.3} & 34.7 & 35.2 & 38.8 \\
    & Prometheus
    & 12.3 & 16.3 & \textbf{17.0} & 12.0 & 11.4
    & 34.7 & 38.6 & 35.9 & 33.7 & \textbf{38.7} \\
    & Mistral
    & 4.81 & \textbf{6.15} & 3.04 & 5.57 & 5.56
    & 45.6 & \textbf{45.7} & 45.4 & 45.6 & 45.4 \\
    \cmidrule(lr){1-12}
    & average
    & 13.0 & \textbf{15.6} & 12.7 & 14.4 & 13.8
    & 34.4 & 36.8 & 32.2 & 35.0 & \textbf{38.0} \\
    \midrule
    (b)
    & Llama
    & 17.1 & 18.3 & \textbf{22.5} & 16.4 & 16.5
    & 9.02 & 13.0 & 14.3 & 7.66 & \textbf{14.4} \\
    $\uparrow$
    & Qwen
    & 26.5 & \textbf{27.0} & 10.8 & 26.3 & 26.1
    & 32.3 & 45.2 & 11.6 & 32.2 & \textbf{49.1} \\
    & Gemma
    & 20.5 & 19.4 & \textbf{21.1} & 19.5 & 20.3
    & 35.8 & 32.9 & \textbf{37.1} & 35.3 & 34.2 \\
    & Prometheus
    & 19.4 & \textbf{21.0} & 20.9 & 19.1 & 17.9
    & 36.6 & \textbf{39.6} & 36.5 & 35.6 & 36.3 \\
    & Mistral
    & 8.74 & \textbf{17.7} & 2.91 & 7.72 & 7.56
    & 47.1 & 47.1 & 47.0 & 47.1 & \textbf{47.2} \\
    \cmidrule(lr){1-12}
    & average
    & 18.4 & \textbf{20.7} & 15.7 & 17.8 & 17.7
    & 32.2 & 35.6 & 29.3 & 31.6 & \textbf{36.2} \\
    \midrule
    (c)
    & Llama
    & 1.59 & 1.92 & 1.70 & 1.57 & \textbf{1.56}
    & 4.86 & 2.37 & 2.39 & 4.25 & \textbf{1.12} \\
    $\downarrow$
    & Qwen
    & \textbf{1.49} & 1.50 & 2.55 & \textbf{1.49} & \textbf{1.49}
    & 1.22 & 1.45 & 3.50 & 1.18 & \textbf{0.840} \\
    & Gemma
    & 1.69 & \textbf{1.50} & 1.83 & 1.55 & 1.70
    & 1.64 & \textbf{1.01} & 1.42 & 1.31 & 1.09 \\
    & Prometheus
    & 2.34 & 2.14 & 2.26 & 2.35 & \textbf{2.10}
    & 2.14 & 1.23 & 1.46 & 1.32 & \textbf{1.17} \\
    & Mistral
    & 1.62 & 2.28 & 3.81 & 1.63 & \textbf{1.60}
    & 0.903 & 0.960 & 0.935 & 0.892 & \textbf{0.872} \\
    \cmidrule(lr){1-12}
    & average
    & 1.75 & 1.87 & 2.43 & 1.72 & \textbf{1.69}
    & 2.15 & 1.40 & 1.94 & 1.79 & \textbf{1.02} \\
    \midrule
    (d)
    & Llama
    & 5.38 & 4.32 & \textbf{2.91} & 6.05 & 7.21
    & 4.26 & 2.05 & \textbf{1.78} & 4.15 & 2.23 \\
    $\downarrow$
    & Qwen
    & 10.6 & 10.6 & \textbf{8.77} & 10.7 & 10.5
    & 4.78 & 4.06 & 7.81 & 5.35 & \textbf{3.61} \\
    & Gemma
    & 4.99 & 10.7 & \textbf{4.10} & 6.88 & 5.19
    & 4.52 & 4.56 & 3.29 & 3.43 & \textbf{3.27} \\
    & Prometheus
    & 3.51 & 2.49 & \textbf{2.41} & 3.65 & 3.55
    & 2.42 & 2.14 & \textbf{1.26} & 2.26 & 1.97 \\
    & Mistral
    & 10.4 & 8.52 & \textbf{7.46} & 10.3 & 11.0
    & 3.27 & 3.00 & \textbf{2.69} & 3.68 & 3.59 \\
    \cmidrule(lr){1-12}
    & average
    & 6.98 & 7.33 & \textbf{5.13} & 7.51 & 7.48
    & 3.85 & 3.16 & 3.37 & 3.78 & \textbf{2.93} \\
    \bottomrule
  \end{tabular}
  \medskip
\end{table}

\begin{table}[tb]
  \centering
  \setlength{\tabcolsep}{2.5pt}
  \caption{Results of the STS and STR tasks. (a) Spearman, (b) Pearson, (c) MSE, and (d) SDD.}
  \label{tab:result-sts-str}
  \begin{tabular}{clcccccccccc}
    \toprule
    & \textbf{LLM}
    & \multicolumn{5}{c}{\textbf{STS}} & \multicolumn{5}{c}{\textbf{STR}} \\
    \cmidrule(lr){3-7}
    \cmidrule(lr){8-12}
    & & Van. & CC & DC & Deb-R & Deb-T & Van. & CC & DC & Deb-R & Deb-T \\
    \midrule
    (a)
    & Llama
    & 7.45 & 31.5 & 2.43 & 14.2 & \textbf{47.0}
    & 26.8 & \textbf{42.9} & 17.2 & 25.6 & 37.1 \\
    $\uparrow$
    & Qwen
    & 81.6 & 81.9 & 81.7 & \textbf{82.2} & 82.1
    & \textbf{73.4} & 72.3 & 73.3 & 73.3 & 73.3 \\
    & Gemma
    & 63.6 & 69.5 & 63.5 & \textbf{70.0} & 69.6
    & 40.1 & \textbf{59.2} & 57.9 & 50.1 & 56.9 \\
    & Prometheus
    & 77.0 & 78.3 & 77.7 & \textbf{79.0} & 78.9
    & \textbf{70.3} & 68.6 & 60.3 & \textbf{70.3} & 70.1 \\
    & Mistral
    & 79.6 & 79.4 & 79.6 & \textbf{79.8} & \textbf{79.8}
    & 70.9 & 70.9 & 70.4 & \textbf{71.1} & 71.0 \\
    \cmidrule(lr){1-12}
    & average
    & 61.9 & 68.1 & 61.0 & 65.0 & \textbf{71.5}
    & 56.3 & \textbf{62.8} & 55.8 & 58.1 & 61.7 \\
    \midrule
    (b)
    & Llama
    & 4.26 & 29.5 & 12.0 & 15.6 & \textbf{48.4}
    & 17.3 & \textbf{40.0} & 13.9 & 20.9 & 23.8 \\
    $\uparrow$
    & Qwen
    & 77.6 & 76.1 & 79.2 & 81.0 & \textbf{81.5}
    & \textbf{71.8} & 63.4 & 71.6 & 71.3 & 71.4 \\
    & Gemma
    & 65.1 & 65.2 & 66.0 & \textbf{68.5} & 67.5
    & 39.0 & 39.0 & \textbf{50.6} & 39.2 & 36.9 \\
    & Prometheus
    & 73.8 & 77.9 & 79.3 & 78.5 & \textbf{80.0}
    & 64.6 & \textbf{66.5} & 56.6 & 64.8 & 63.5 \\
    & Mistral
    & 76.2 & 78.8 & 77.2 & \textbf{79.9} & 79.7
    & 67.8 & \textbf{71.6} & 68.3 & 67.6 & 67.7 \\
    \cmidrule(lr){1-12}
    & average
    & 59.4 & 65.5 & 62.7 & 64.7 & \textbf{71.4}
    & 52.1 & \textbf{56.1} & 52.2 & 52.8 & 52.7 \\
    \midrule
    (c)
    & Llama
    & 9.12 & \textbf{2.21} & 6.61 & 3.37 & 2.25
    & 6.99 & \textbf{0.661} & 1.43 & 6.27 & 1.31 \\
    $\downarrow$
    & Qwen
    & 0.976 & 1.02 & \textbf{0.926} & 0.935 & 0.934
    & 0.862 & \textbf{0.526} & 0.781 & 0.661 & 0.746 \\
    & Gemma
    & 1.85 & 1.84 & 1.79 & \textbf{1.41} & 1.50
    & 2.99 & 1.88 & \textbf{1.48} & 2.16 & 2.02 \\
    & Prometheus
    & 1.55 & \textbf{1.10} & 1.12 & 1.45 & 1.73
    & 2.16 & \textbf{0.508} & 0.778 & 1.53 & 0.971 \\
    & Mistral
    & 1.55 & \textbf{1.29} & 1.36 & 1.36 & 1.44
    & 1.31 & 1.26 & 1.02 & 1.23 & \textbf{0.996} \\
    \cmidrule(lr){1-12}
    & average
    & 3.01 & \textbf{1.49} & 2.36 & 1.71 & 1.57
    & 2.86 & \textbf{0.967} & 1.10 & 2.37 & 1.21 \\
    \midrule
    (d)
    & Llama
    & 11.1 & 2.25 & 3.10 & \textbf{2.17} & 2.91
    & 7.59 & \textbf{1.04} & 1.27 & 7.51 & 2.54 \\
    $\downarrow$
    & Qwen
    & 7.33 & 7.72 & 6.45 & 1.92 & \textbf{1.67}
    & 3.27 & \textbf{3.14} & 3.17 & 3.20 & 3.22 \\
    & Gemma
    & 4.97 & 8.69 & 4.24 & \textbf{1.54} & 1.64
    & 6.82 & 7.45 & \textbf{3.44} & 6.62 & 7.08 \\
    & Prometheus
    & 1.71 & 2.09 & \textbf{1.19} & \textbf{1.19} & 1.25
    & 3.11 & \textbf{0.922} & 0.933 & 2.82 & 2.66 \\
    & Mistral
    & 7.06 & 7.18 & 6.11 & 2.18 & \textbf{1.79}
    & 6.15 & 6.12 & \textbf{4.55} & 6.13 & 6.00 \\
    \cmidrule(lr){1-12}
    & average
    & 6.44 & 5.59 & 4.22 & \textbf{1.80} & 1.85
    & 5.39 & 3.73 & \textbf{2.67} & 5.25 & 4.30 \\
    \bottomrule
  \end{tabular}
\end{table}

Tables~\ref{tab:result-llla-summ} and \ref{tab:result-sts-str} show the results of the debiasing methods using five different LLM evaluators. 
On the one hand, for the Summ, STS, and STR tasks, the proposed methods (Deb-R and Deb-T) outperform the vanilla in most cases.
On the other hand, for the LLM alignment task, the proposed methods achieve slightly better MSE than the vanilla method, but not for other criteria, indicating that the bias of LLMs is not sufficiently mitigated.
The reason for it will be discussed in Subsection~\ref{subsec:error_analysis}.
\par
Hereafter, we discuss the results of the tasks except for the LLM alignment.
Our methods achieve superior performance to the previous debiasing methods, DC for all tasks, and CC for all tasks except for STR. Especially, Deb-T with Llama improves Spearman's correlation by 15.5 and 44.6 points over CC and DC, respectively, for the STS task. 
It indicates that measuring the inherent bias of LLMs in generating numerical tokens via random number generation is effective in mitigating the scoring bias. \par
Deb-T outperforms Deb-R in most cases. 
In particular, Deb-T with Llama achieves significantly higher Spearman's and Pearson's correlations in the STS task. 
These results support the idea that the latent number bias of LLMs is task-dependent and that adding instructions for the downstream task is effective for measuring it more precisely.

\subsection{Evaluation of Prompt Selection}
\label{subsec:effect_on_prompt}
As described in Subsection~\ref{subsec:calculating_latent_bias}, 100 prompts for random number generation are generated, and the top $T_p$ prompts with low perplexity are chosen.
The parameter $T_p$ was set to 10 in the previous experiments, but it may affect debiasing performance.
To evaluate the potential impact of selecting reliable prompts on measuring latent number bias in LLMs, the methods Deb-R with $T_p=10$ and $T_p=100$ (without prompt selection) are compared.\par
Table~\ref{tab:debr-10-vs-100-all} shows Spearman’s correlation of two methods.
Deb-R$_{100}$ ($T_p=100$) outperforms Deb-R$_{10}$ ($T_p=10$) overall; however, the differences between them are small in most cases.
These results suggest that the effectiveness of the prompt selection on random number generation is limited.

\begin{table}[tb]
  \centering
  \caption{Spearman's correlation of Deb-R$_{10}$ and Deb-R$_{100}$. Bold indicates a better method.}
  \label{tab:debr-10-vs-100-all}
  \small
  \setlength{\tabcolsep}{4pt}
  \begin{tabular}{lcccccccc}
    \toprule
    LLM
    & \multicolumn{2}{c}{LLM-Align}
    & \multicolumn{2}{c}{Summ}
    & \multicolumn{2}{c}{STS}
    & \multicolumn{2}{c}{STR} \\
    \cmidrule(lr){2-3}
    \cmidrule(lr){4-5}
    \cmidrule(lr){6-7}
    \cmidrule(lr){8-9}
    & R$_{10}$ & R$_{100}$
    & R$_{10}$ & R$_{100}$
    & R$_{10}$ & R$_{100}$
    & R$_{10}$ & R$_{100}$ \\
    \midrule

    Llama
    & \textbf{14.9} & 14.3
    & 13.0 & 13.0
    & 14.2 & \textbf{23.1}
    & 25.6 & 25.6 \\
    Qwen
    & 23.6 & \textbf{23.8}
    & \textbf{47.6} & 47.3
    & 82.2 & 82.2
    & 73.3 & \textbf{73.4} \\
    Gemma
    & 15.7 & \textbf{17.2}
    & 35.2 & \textbf{35.3}
    & 70.0 & \textbf{70.3}
    & 50.1 & \textbf{55.3} \\
    Prometheus
    & 12.0 & \textbf{12.3}
    & 33.7 & \textbf{33.8}
    & 79.0 & 79.0
    & \textbf{70.3} & 70.2 \\
    Mistral
    & 5.57 & \textbf{5.66}
    & 45.6 & 45.6
    & 79.8 & \textbf{80.0}
    & 71.1 & 71.1 \\
    \cmidrule(lr){1-9}
    average
    & 14.4 & \textbf{14.7}
    & 35.0 & 35.0
    & 65.0 & \textbf{66.9}
    & 58.1 & \textbf{59.1} \\
    \bottomrule
  \end{tabular}
\end{table}

\subsection{Evaluation of Search Range Expansion of Hyperparameter}
\label{subsec:strength_of_lambda}
In previous experiments, the hyperparameter $\lambda$, which controls the intensity of debiasing, was optimized over the range $0.1 \leq \lambda \leq 1$.
It is expected that extending the search range could improve the performance of LLM evaluators.
Therefore, we implement \mbox{Deb-R$_{\lambda \le 10}$} and \mbox{Deb-T$_{\lambda \le 10}$}, which are variants of Deb-R and Deb-T, respectively, in which $\lambda$ is optimized over the range $0.1 \leq \lambda \leq 10$.
Table~\ref{tab:lambda-range-all-models} shows Spearman's correlation for these settings as well as the original settings (denoted as \mbox{Deb-R$_{\lambda \le 1}$} and \mbox{Deb-T$_{\lambda \le 1}$}).\par
Although \mbox{Deb-R$_{\lambda \le 10}$} is significantly better than \mbox{Deb-R$_{\lambda \le 1}$} in some cases, e.g., the Llama-based model for the STS and STR tasks, unexpectedly, the extension of the search range does not improve performance in many cases.
A similar trend is observed for \mbox{Deb-T$_{\lambda \le 10}$}.
It indicates a gap between the development and test sets; the best $\lambda$ optimized on the development set does not always yield the best performance on the test set.
Optimizing $\lambda$ is important, yet it poses a considerable challenge.

\begin{table}[!t]
  \centering
   \caption{Spearman's correlation of methods with different parameter optimization. Bold indicates \mbox{Deb-R$_{\lambda<10}$}/Deb-T$_{\lambda<10}$ achieves better performance than Deb-R$_{\lambda<1}$/Deb-T$_{\lambda<1}$, while underline indicates the performance of these two methods is equal.}
  \label{tab:lambda-range-all-models}

  \newcommand{\und}[1]{\underline{#1}}
  \setlength{\tabcolsep}{3pt}
  \begin{tabular}{llcccccccc}
    \toprule
    Method & LLM
    & \multicolumn{2}{c}{LLM-Align}
    & \multicolumn{2}{c}{Summ}
    & \multicolumn{2}{c}{STS}
    & \multicolumn{2}{c}{STR} \\
    \cmidrule(lr){3-4}
    \cmidrule(lr){5-6}
    \cmidrule(lr){7-8}
    \cmidrule(lr){9-10}
    & 
    & $\lambda \leq 1$ & $\lambda \leq 10$
    & $\lambda \leq 1$ & $\lambda \leq 10$
    & $\lambda \leq 1$ & $\lambda \leq 10$
    & $\lambda \leq 1$ & $\lambda \leq 10$ \\
    \midrule

    Deb-R
    & Llama
    & 14.9 & \textbf{17.7} 
    & 13.0 & 10.2 
    & 14.2 & \textbf{41.9} 
    & 25.6 & \textbf{38.4} \\
    & Qwen
    & 23.6 & 22.7 
    & 47.6 & 47.2 
    & \und{82.2} & \und{82.2} 
    & \und{73.3} & \und{73.3} \\ 
    & Gemma
    & 15.7 & \textbf{16.6} 
    & 35.2 & 30.0 
    & \und{70.0} & \und{70.0} 
    & 50.1 & \textbf{58.5} \\
    & Prometheus
    & 12.0 & 11.7 
    & 33.7 & 32.7 
    & \und{79.0} & \und{79.0} 
    & \und{70.3} & \und{70.3} \\ 
    & Mistral
    & 5.57 & \textbf{5.73} 
    & 45.6 & 39.4 
    & \und{79.8} & \und{79.8} 
    & 71.1 & \textbf{71.2} \\
    \cmidrule(lr){2-10}
    & average
    & 14.4 & \textbf{14.9} 
    & 35.0 & 31.9 
    & 65.0 & \textbf{70.6} 
    & 58.1 & \textbf{62.3} \\
    \midrule

    Deb-T
    & Llama
    & 14.9 & \textbf{20.2}
    & 18.0 & 13.1
    & \und{47.0} & \und{47.0}
    & 37.1 & \textbf{38.5} \\
    & Qwen
    & 23.6 & 22.3
    & 49.0 & 30.0
    & \und{82.1} & \und{82.1}
    & \und{73.3} & \und{73.3} \\
    & Gemma
    & 13.6 & 13.0
    & 38.8 & \textbf{42.5}
    & \und{69.6} & \und{69.6}
    & 56.9 & \textbf{59.1} \\
    & Prometheus
    & 11.4 & \textbf{15.8}
    & 38.7 & 27.7 
    & \und{78.9} & \und{78.9}
    & \und{70.1} & \und{70.1} \\
    & Mistral
    & 5.56 & 5.22
    & 45.4 & 44.0 
    & \und{79.8} & \und{79.8}
    & \und{71.0} & \und{71.0} \\
    \cmidrule(lr){2-10}
    & average
    & 13.8 & \textbf{15.3}
    & 38.0 & 31.4 
    & \und{71.5} & \und{71.5}
    & 61.7 & \textbf{62.4} \\
    \bottomrule
  \end{tabular}
\end{table}

\section{Analysis}
\label{sec:analysis}
\subsection{Analysis on observed latent number bias}
Figure~\ref{fig:random_probs_all} shows the latent number bias of five LLMs using four different methods: Deb-R$_{10}$, Deb-T$_{10}$, Deb-R$_{100}$, and Deb-T$_{100}$.
First, we discuss the task-agnostic latent number bias, which can be observed by the method Deb-R.
Different latent number biases are observed across LLMs.
Llama and Prometheus exhibit a preference for the highest scores.
Gemma shows a bias toward the lowest and highest scores.
Qwen and Mistral tend to generate the number `1'.
The range of scores in random number generation also influences the latent number bias. 
For example, Qwen and Mistral exhibit different latent number biases when the scoring range is [1,5] (i.e., the Summ task) compared to other scoring ranges.

Adding the task definition to the prompt for random number generation significantly changes the latent number bias, as confirmed by the comparison between Deb-R (blue lines) and Deb-T (yellow lines).
Therefore, it is essential to measure and mitigate latent number bias for a specific downstream task.
This is supported by the fact that Deb-T outperformed Deb-R in Tables 1 and 2.

Besides, the latent number bias is nearly identical between Deb-R$_{10}$ (or Deb-T$_{10}$), indicated by a solid line, and Deb-R$_{100}$ (or Deb-T$_{100}$), indicated by a dotted line.
The number of prompts used for random number generation ($T_p=10$ vs. $T_p=100$) is not a significant factor in the scoring bias of LLMs.

\subsection{Analysis of Poor Performance in LLM Alignment Task}
\label{subsec:error_analysis}
\begin{figure}[!t]
    \centering
    \begin{minipage}{0.8\linewidth}
      \centering
      \includegraphics[width=\linewidth]{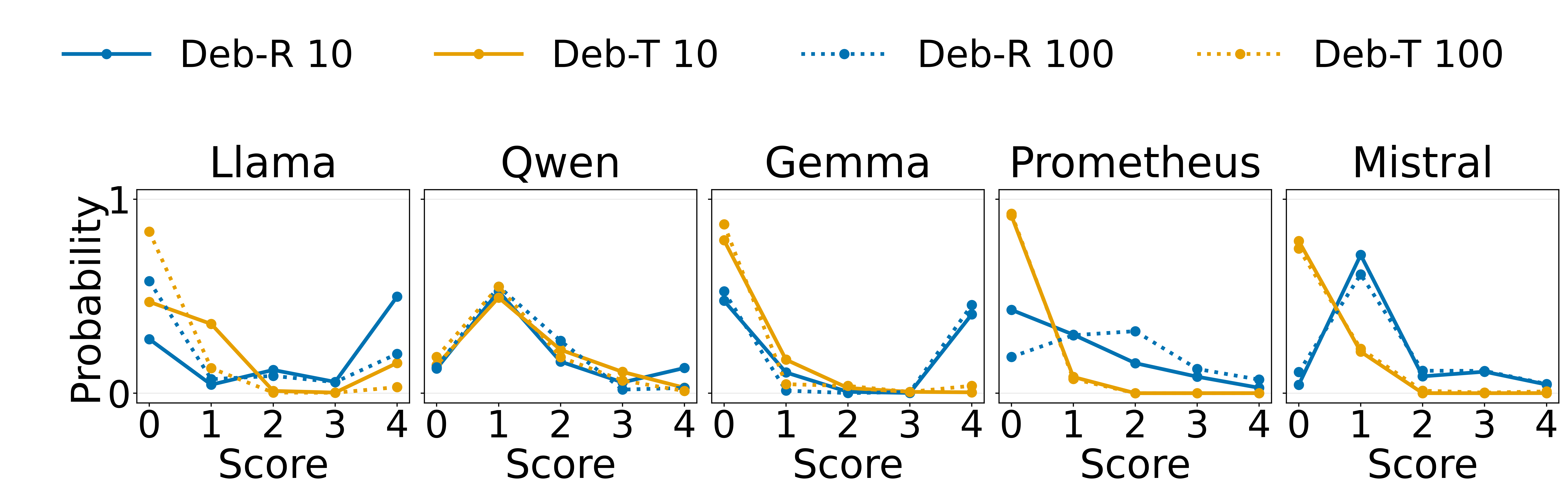}
      \subcaption{LLM-Align}\label{fig:random_probs_llm}

      \includegraphics[width=\linewidth]{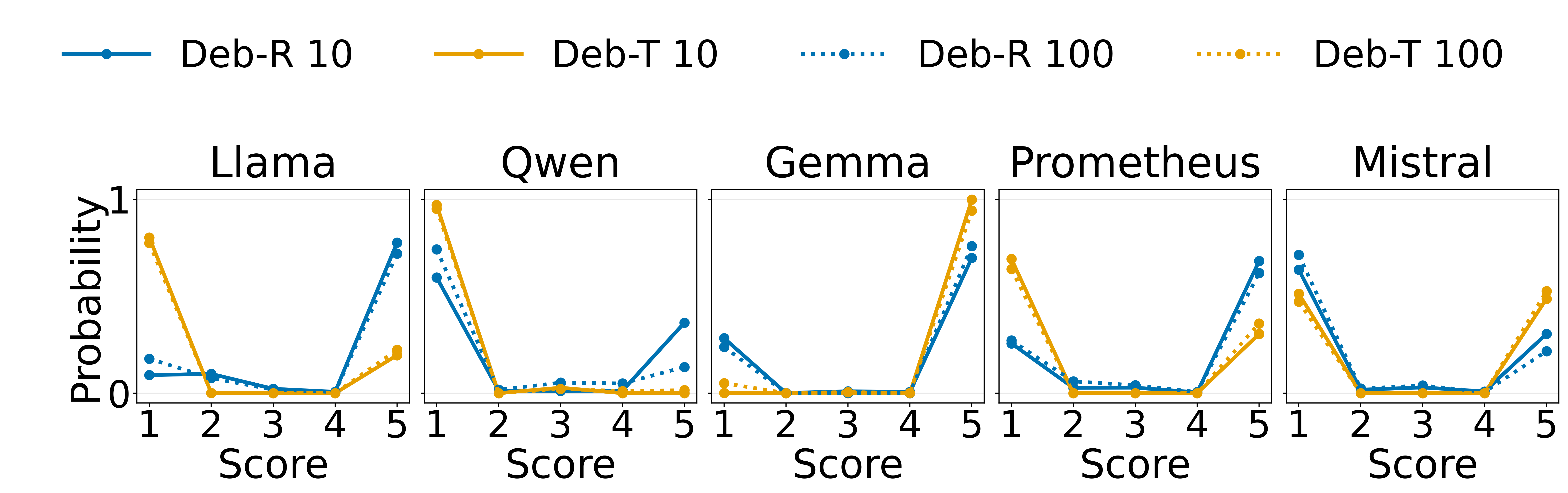}
      \subcaption{Summ}\label{fig:random_probs_summ}

      \includegraphics[width=\linewidth]{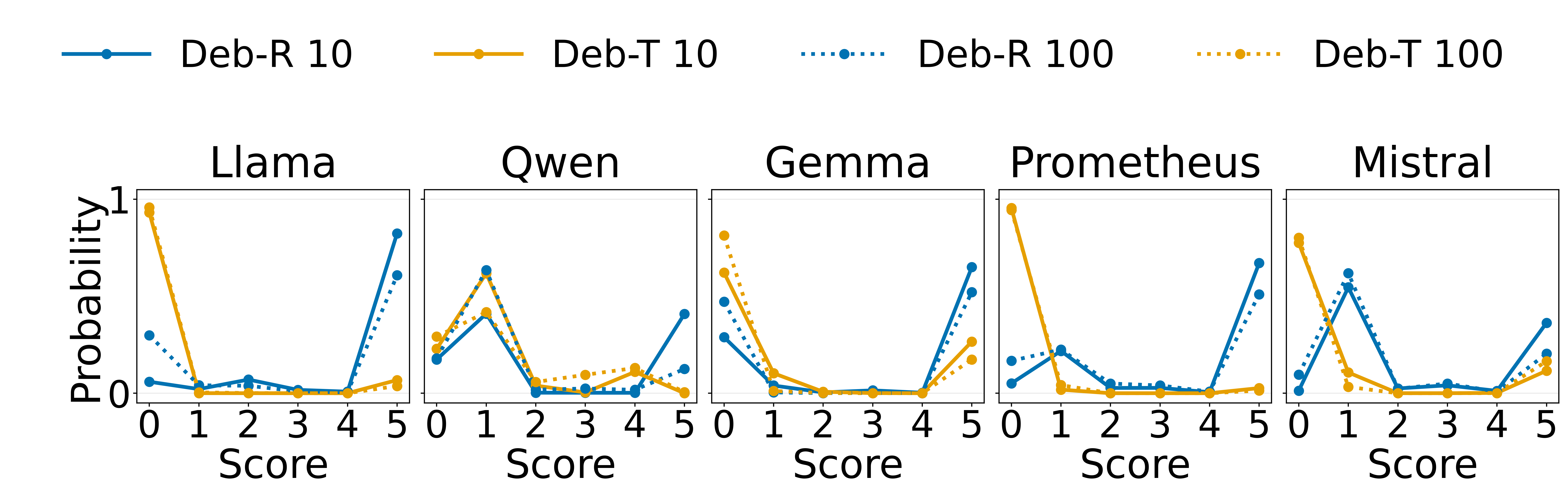}
      \subcaption{STS}\label{fig:random_probs_sts}

      \includegraphics[width=\linewidth]{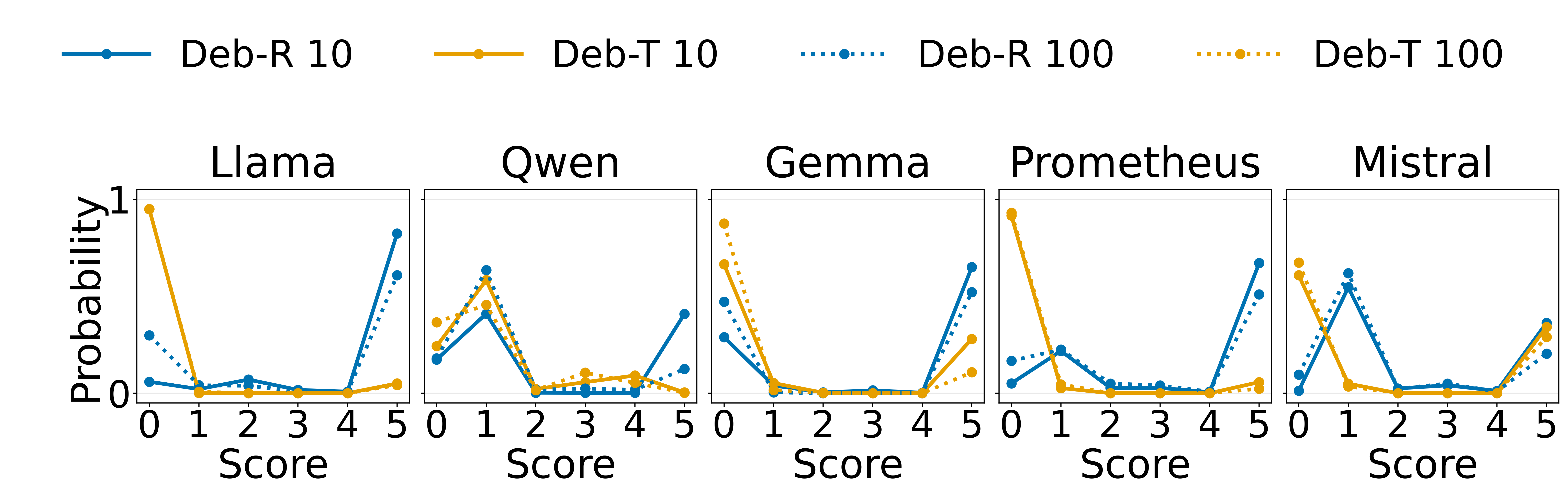}
      \subcaption{STR}\label{fig:random_probs_str}
    \end{minipage}
    
    \caption{Latent number bias of LLMs.}
    \label{fig:random_probs_all}
\end{figure}

\begin{figure}[htbp]
    \centering
    \includegraphics[width=\columnwidth]{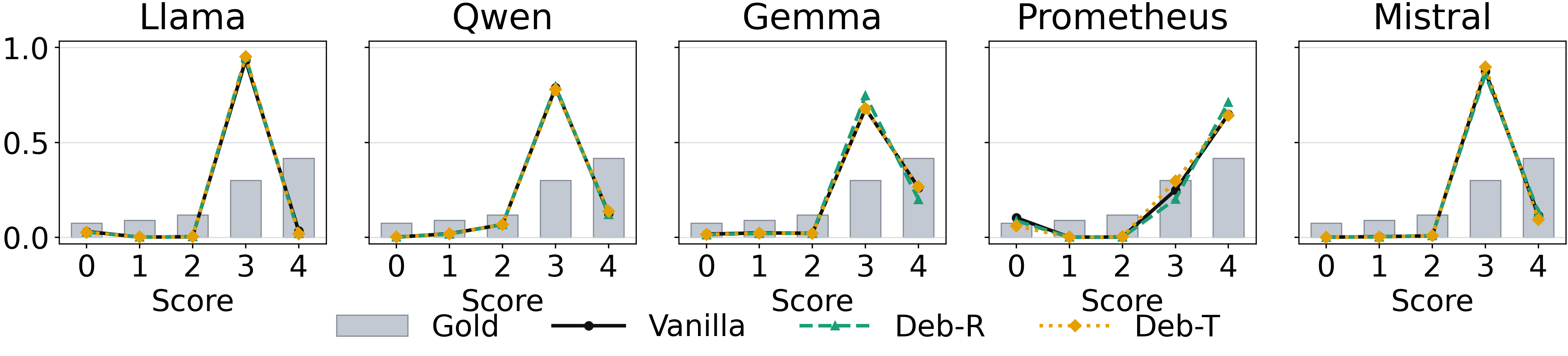}
    \caption{Proportion of ground-truth and predicted scores for LLM Alignment task.}
    \label{fig:dist_llm_align}
    \medskip
\end{figure}

The poor performance of the proposed debiasing method on the LLM alignment task, as shown in Table~\ref{tab:result-llla-summ}, is due to the limited capacity of the pre-trained LLMs.
Fig.~\ref{fig:dist_llm_align} shows the proportions of the scores predicted by the LLM evaluators and the ground-truth scores.
The lines indicate the average generation probability for each predicted number from the vanilla, Deb-R, and Deb-T models, while the bar chart shows the proportion of ground-truth scores.
As the score increases, the number of samples in the test set increases. 
However, vanilla LLMs, except for Prometheus, predict the score `3' (not the max score `4') for most samples.
This deviation was not sufficiently revised by our debiasing method.
The latent number bias shown in Figure~\ref{fig:random_probs_llm} exhibits low bias toward the score `3'. Thus, the prediction probability of the score `3' is further increased by debiasing.
Therefore, debiasing cannot have a positive impact on LLM evaluators, resulting in almost equivalent performance before and after debiasing.

\FloatBarrier

\section{Conclusion}
\label{sec:conclusion}
We proposed a debiasing method for LLM-as-a-Judge to mitigate scoring bias.
The latent number bias of an LLM was measured by random number generation, and the generation probability of the number token was increased (or decreased) when it had a low (or high) inherent preference in the observed latent number bias.
The results of the experiments on four evaluation tasks using five LLMs showed the superiority of the proposed method over existing methods.
In addition, the effectiveness of the variants of the proposed methods was investigated, including adding the task definition to the prompt for random number generation, changing the number of prompts for random number generation, and optimizing the weight parameter to control the degree of debiasing. \par
This study leaves several issues to be addressed in the future.
First, the effectiveness of our method should be assessed across a wider range of evaluation tasks to evaluate its robustness.
Second, instead of using a naive zero-shot prompt in an LLM evaluator, more sophisticated prompts, such as few-shot and chain-of-thought prompts, should be employed.
Third, it is worth investigating to determine the hyperparameter $\lambda$ for individual samples.
We confirmed that the vanilla LLMs can precisely evaluate the quality of several samples, for which debiasing is unnecessary.
Setting the $\lambda$ to zero for these samples would improve the overall performance of LLM evaluators.
Finally, larger LLMs should be used in the experiment to examine the relationship between the number of parameters in an LLM and the effectiveness of debiasing.

\FloatBarrier

\appendix
\section{Prompt}
\label{sec:prompt}
Figures~\ref{fig:prompt-llm_align}--\ref{fig:prompt-str} show the prompts for evaluation of LLM alignment, evaluation of summarization, STS, and STR tasks.

\begin{figure}[tb]
\centering

\newlength{\widthofprompt}
\setlength{\widthofprompt}{0.48\linewidth}

\medskip

\begin{minipage}[t]{\widthofprompt}
  \begin{promptbox}
Evaluation Task: Assess Helpfulness \\ 

Definition: \\
"Overall helpfulness of the response to the prompt." \\

Please evaluate the response below based on this definition. \\

Scoring Guide: \\
- 4 (Extremely helpful): The response perfectly and comprehensively addresses the user's prompt. \\
- 3 (Helpful): The response successfully addresses the prompt but could be slightly more comprehensive. \\
- 2 (Moderately helpful): The response partially answers the prompt but has significant room for improvement. \\
- 1 (Slightly helpful): The response makes an attempt but is largely unhelpful or misses the main point. \\
- 0 (Not helpful): The response is completely irrelevant or fails to address the prompt. \\

Prompt:  \\
\{\{prompt\}\} \\

Response:\\
\{\{response\}\} \\ 

Helpfulness (0-4): 
  \end{promptbox}

  \smallskip

  \caption{Prompt for the LLM-Align task.}
  \label{fig:prompt-llm_align}
\end{minipage}
\hfill
\begin{minipage}[t]{\widthofprompt}
  \begin{promptbox}
You will be given one summary written for a news article.\\ 

Your task is to rate the summary on one metric.\\ 

Please make sure you read and understand these instructions carefully. Please keep this document open while reviewing, and refer to it as needed.\\ 

Evaluation Criteria:\\ 

Coherence (1-5) - the collective quality of all sentences. We align this dimension with the DUC quality question of structure and coherence whereby "the summary should be well-structured and well-organized. The summary should not just be a heap of related information, but should build from sentence to a coherent body of information about a topic."\\

Source Text:\\ 
\{\{Source\}\}\\ 

Summary:\\ 
\{\{Summary\}\}\\

Evaluation Form (scores ONLY):\\ 

- Coherence:
  \end{promptbox}

  \smallskip

  \caption{Prompt for the Summ task.}
  \label{fig:prompt-summeval}
\end{minipage}

\begin{minipage}[t]{\widthofprompt}
  \begin{promptbox}
Task:\\ 

You will evaluate the semantic textual similarity (STS) between two sentences.\\ 
Please score the similarity between the following two sentences on a scale from 0 to 5.\\

Sentence 1: \{\{Text 1\}\}\\ 
Sentence 2: \{\{Text 2\}\}\\ 

Score:
  \end{promptbox}

  \smallskip

  \caption{Prompt for the STS task.}
  \label{fig:prompt-stsb}
\end{minipage}
\hfill
\begin{minipage}[t]{\widthofprompt}
  \begin{promptbox}
Task: \\ 

You will solve the Semantic Relatedness (SemRel 2024) task. \\
Please score the Semantic Relatedness between the following two sentences on a scale from 0 to 5. \\

Sentence 1: \{\{Text 1\}\} \\  
Sentence 2: \{\{Text 2\}\} \\ 
  
Score: 
  \end{promptbox}

  \smallskip

  \caption{Prompt for the STR task.}
  \label{fig:prompt-str}
\end{minipage}

\end{figure}

\FloatBarrier
%
%
%

\bibliographystyle{splncs04}
\bibliography{references}



\end{document}